\documentclass{article}
\usepackage{iclr2025_conference,times}

\usepackage{wrapfig}
\usepackage{lipsum}
\usepackage{wrapfig}

\usepackage{amsmath,amsfonts,bm}
\usepackage{multicol}
\def\eqref#1{equation~\ref{#1}}
\def\1{\bm{1}}

\DeclareMathAlphabet{\mathsfit}{\encodingdefault}{\sfdefault}{m}{sl}
\SetMathAlphabet{\mathsfit}{bold}{\encodingdefault}{\sfdefault}{bx}{n}

\usepackage{booktabs}
\usepackage{multirow}
\usepackage{amssymb}
\usepackage{graphicx, subfigure}

\usepackage{colortbl}
\usepackage{xcolor}

\usepackage{hyperref}
\usepackage{url}
\usepackage{multicol}
\usepackage{aliascnt}
\usepackage{adjustbox}
\usepackage{siunitx}
\usepackage{booktabs}
\usepackage{multirow} 
\usepackage{enumitem}
\usepackage{booktabs}
\usepackage{amssymb}
\usepackage{amsmath}
\usepackage{graphicx}
\usepackage{longtable}
\usepackage{graphicx}
\usepackage{subcaption}
\usepackage{calc}
\usepackage[most]{tcolorbox}
\usepackage{listings}
\usepackage{latexsym}
\usepackage{dsfont}
\usepackage{mathtools}
\usepackage{makecell}

\usepackage[ruled]{algorithm2e}

\definecolor{uclablue}{rgb}{0.15, 0.45, 0.68}
\hypersetup{
    breaklinks,
    citecolor=uclablue,
    colorlinks=true,
}

\definecolor{lightgreen}{RGB}{0,150,0}
\definecolor{myred}{RGB}{200,0,0}

\usepackage{tabularx}
\robustify\bfseries
\usepackage{subcaption}
\usepackage{cleveref}

\tcbuselibrary{listings,skins,breakable}

\newtcolorbox{prompt}[1][]{enhanced,
  breakable,
  colback=white,
  colframe=black,
  coltitle=black,
  colbacktitle=gray!20,
  fonttitle=\bfseries,
  title=Prompt,
  #1}

\newtcolorbox{remark}[1][]{enhanced,
  breakable,
  colback=violet!6,
  colframe=violet!50!black,
  coltitle=black,
  colbacktitle=violet!18,
  fonttitle=\bfseries,
  title=Remark,
  #1}

\definecolor{linkColor}{rgb}{0.2,0.4,0.6}
\definecolor{myblue}{HTML}{0379AC}
\definecolor{myred}{HTML}{A50E50}
\definecolor{myorange}{RGB}{238, 133, 74}
\definecolor{latentcolor}{named}{cyan}
\definecolor{normalcolor}{RGB}{0, 0, 0}
\usepackage{marvosym}

\usepackage{xcolor}

\title{TMP: Tree-structured Mixed-policy Pruning for Large-scale Image Generation and Editing}

\author{
Peizhen Zhang, Yang Li, Xunsong Li, Songtao Liu\thanks{Corresponding author}, Zewen Liu, Qiangqiang Hu, Guotong Guo, Jupeng Ding, Yifu Sun, coopersli, Jian Zhang, Zhao Zhong\thanks{Project leader}, Liefeng Bo \\ 
 \textbf{Multimodal Model Department, Tencent}\\
}

\begin{document}
\maketitle
\let\oldthefootnote\thefootnote

\let\thefootnote\oldthefootnote

\begin{abstract}
Modern image generation model rapidly grows their sizes to meet high-fidelity image synthesis. However, they gradually become unaffordable for their enormous parameter consumption and computation budget that lead to massive resources requirement and gpu memory footprint. In this paper, we propose \textbf{TMP}, the first \textbf{T}ree-structured \textbf{M}ixed-policy \textbf{P}runing framework that generalizes prevalent image tasks (T2I and TI2I) and architectures (Mixture-of-Experts (MoE) and Diffusion transformer (DiT)). It could be applied to the step-distilled models and contribute as the last stage. We perform experiments upon current open-sourced SOTA HunyuanImage-3.0 instruct and a popular efficient model Z-Image turbo. The proposed pruning framework manages to compress HunyuanImage 3.0 from 80B to 20B parameters at \textbf{75\%} reduction ratio, sacrificing limited generation quality. We also optimize to enable the inference of the pruned 20B version of HunyuanImage 3.0 on a single 24GB 4090 GPU by engineering skills. The inference script and model weight have been integrated into the existing HunyuanImage3.0 open-source github\footnote{ \url{https://github.com/Tencent-Hunyuan/HunyuanImage-3.0}.}~\looseness=-1 and huggingface\footnote{ \url{https://huggingface.co/tencent/HunyuanImage-3.0}.}~\looseness=-1 repository. Besides, we prove the efficacy of TMP by compressing Z-Image turbo from 6B to 4B (\textbf{33\%} reduction) with negligible degradation. 
\end{abstract}

\begin{figure*}[t]
\centering

\begin{tabular}{ccc}
\textbf{Source Image}
&
\textbf{Original (80B)}
&
\textbf{Pruned (20B)}
\end{tabular}

\vspace{1mm}

{\footnotesize
Have the man turn around to face the car, with his back to the camera.
}

\vspace{1mm}

\includegraphics[width=0.325\textwidth]
{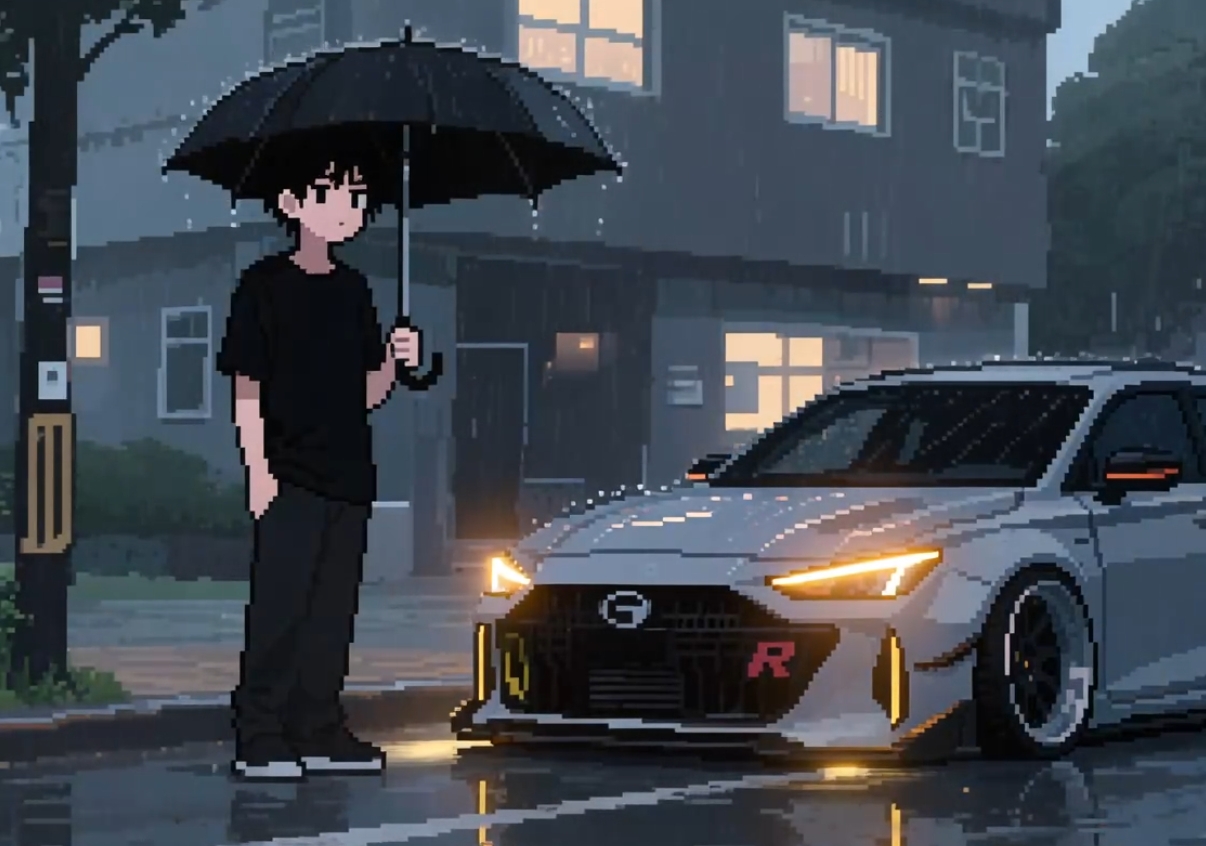}
\hfill
\includegraphics[width=0.325\textwidth]
{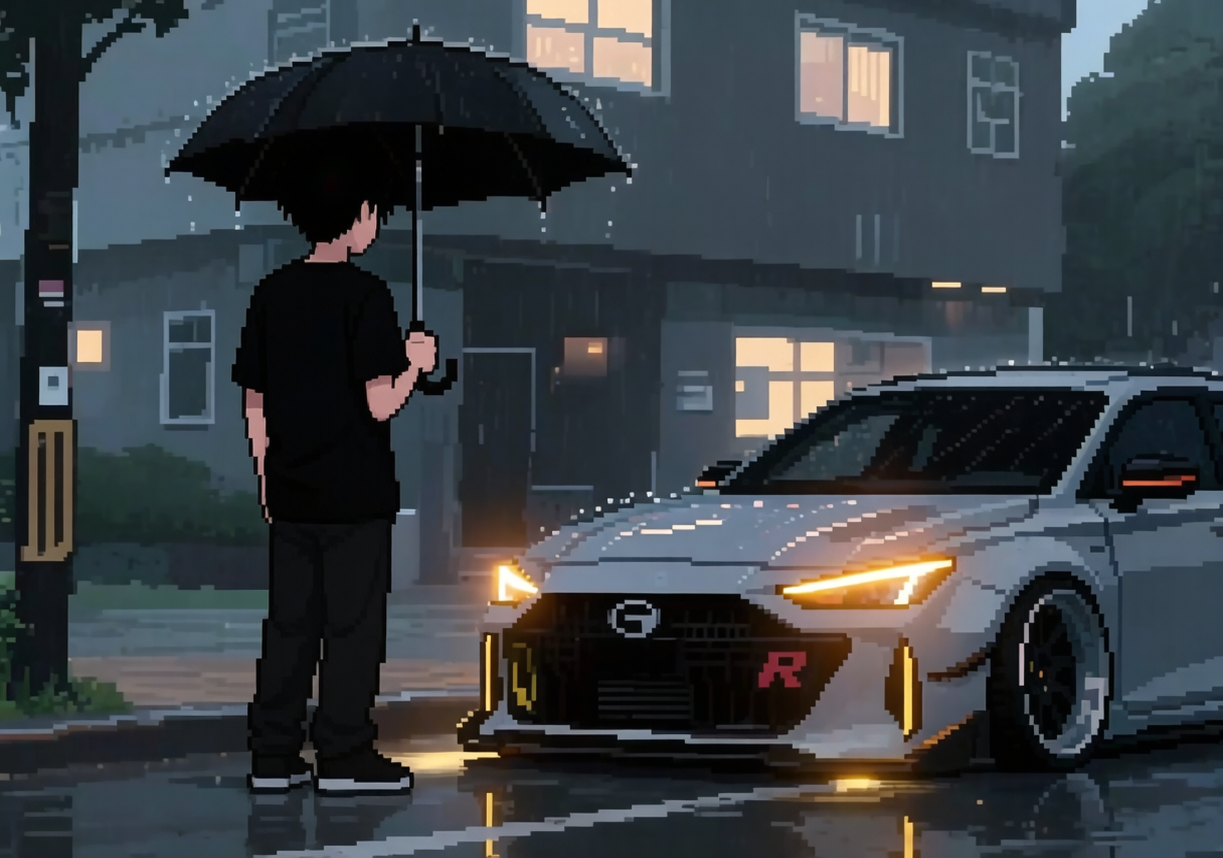}
\hfill
\includegraphics[width=0.325\textwidth]
{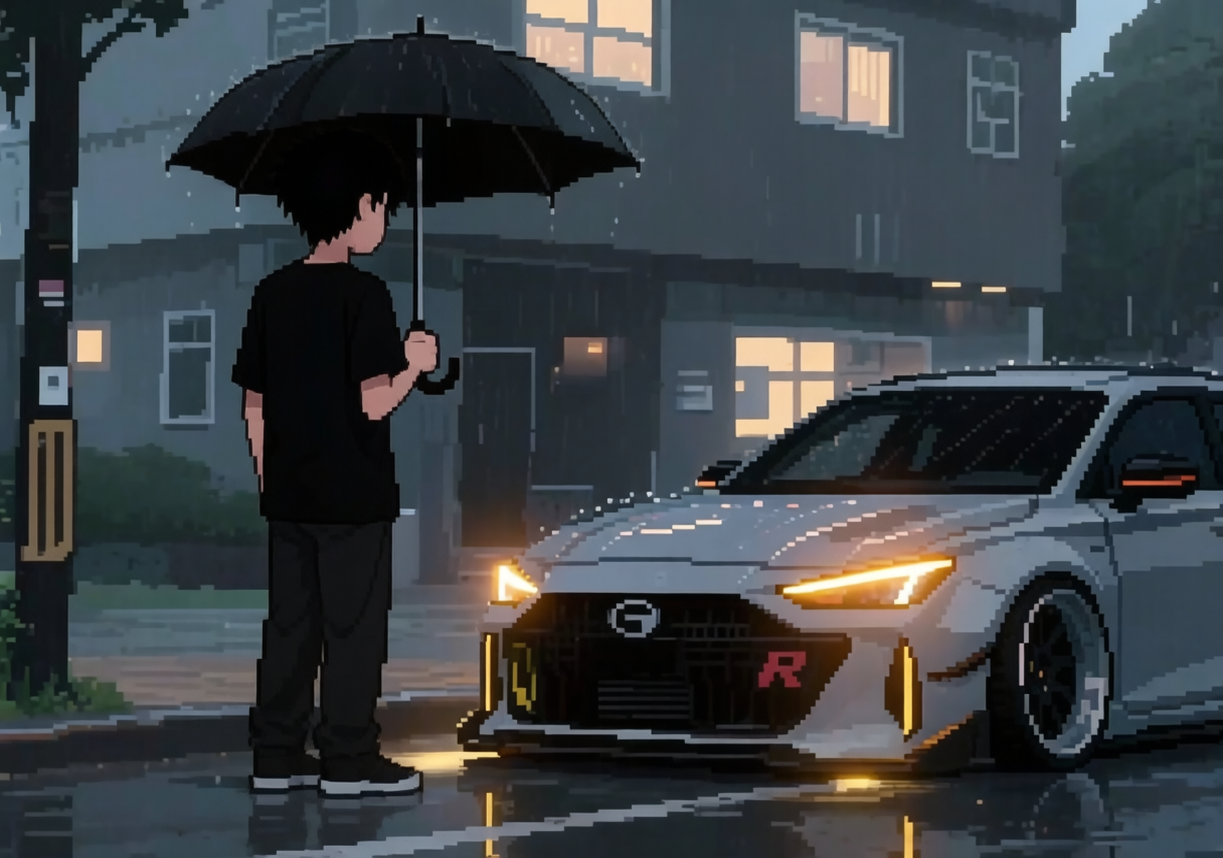}

\vspace{2mm}

{\footnotesize
Generate a young man in the reference style, wearing a purple hat and
red-rimmed glasses, looking toward the left side of the frame.
}

\vspace{1mm}

\includegraphics[width=0.325\textwidth]
{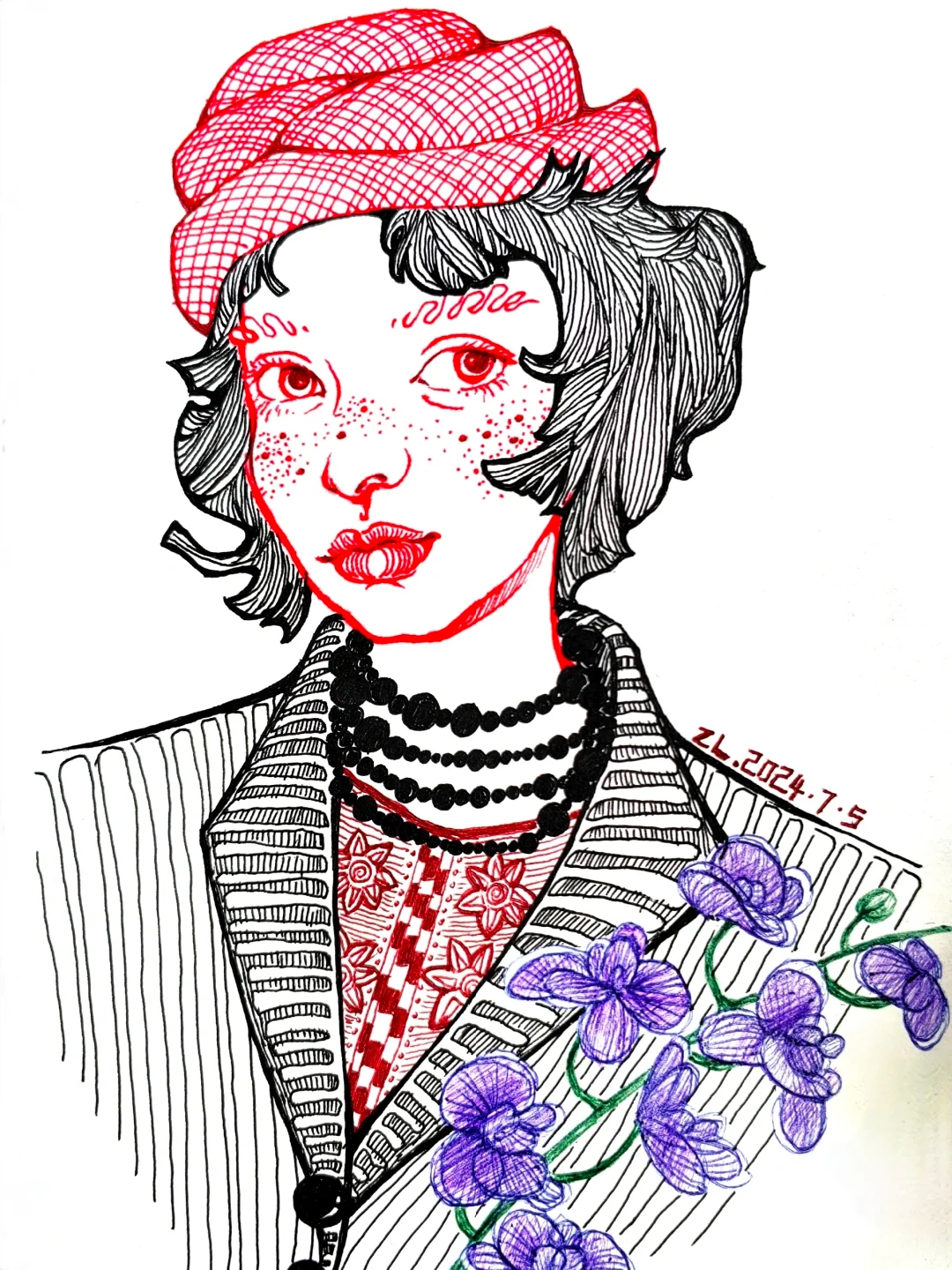}
\hfill
\includegraphics[width=0.325\textwidth]
{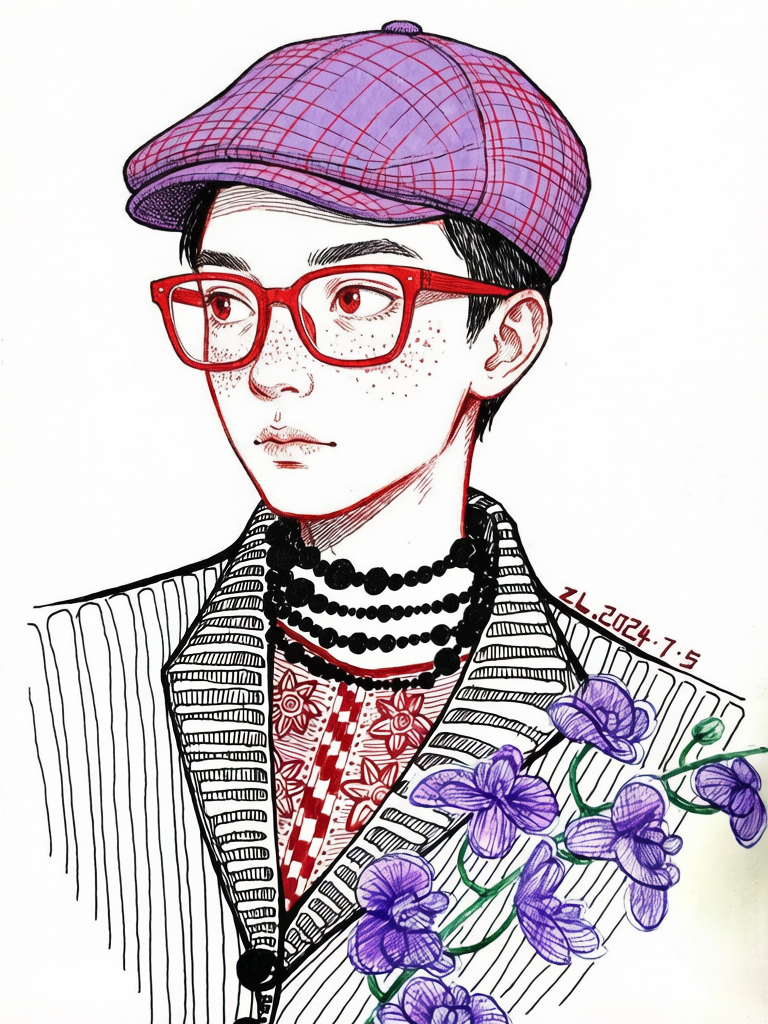}
\hfill
\includegraphics[width=0.325\textwidth]
{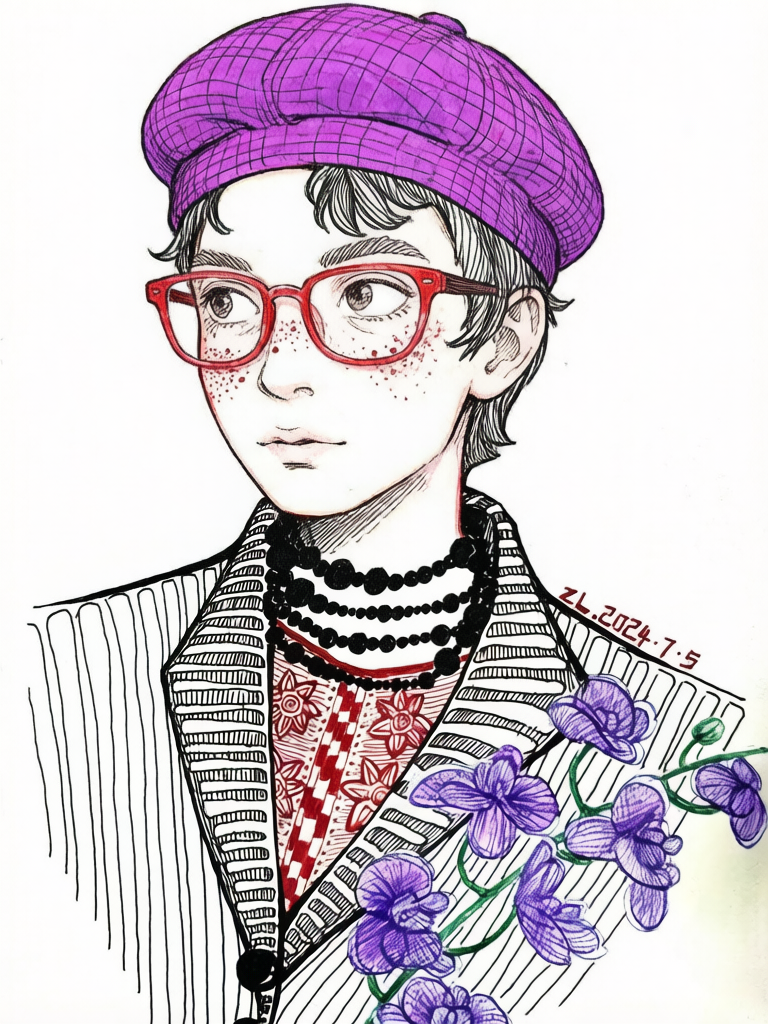}

\vspace{-1mm}

\caption{
Single-reference image editing examples comparing the original
HunyuanImage-3.0 80B model and our 20B pruned model.
Despite a 4$\times$ reduction in model size, the pruned model preserves
editing fidelity and visual quality across diverse editing instructions.
}
\label{fig:single_ref_case}

\vspace{-3mm}

\end{figure*}

\begin{figure*}[t]
\centering

\begin{tabular}{cccc}
\textbf{Reference 1}
&
\textbf{Reference 2}
&
\textbf{Original (80B)}
&
\textbf{Pruned (20B)}
\end{tabular}

\vspace{1mm}

{\footnotesize
Transform the style of a street-view image into the style of a sky image.
}

\vspace{1mm}

\includegraphics[width=0.24\textwidth]
{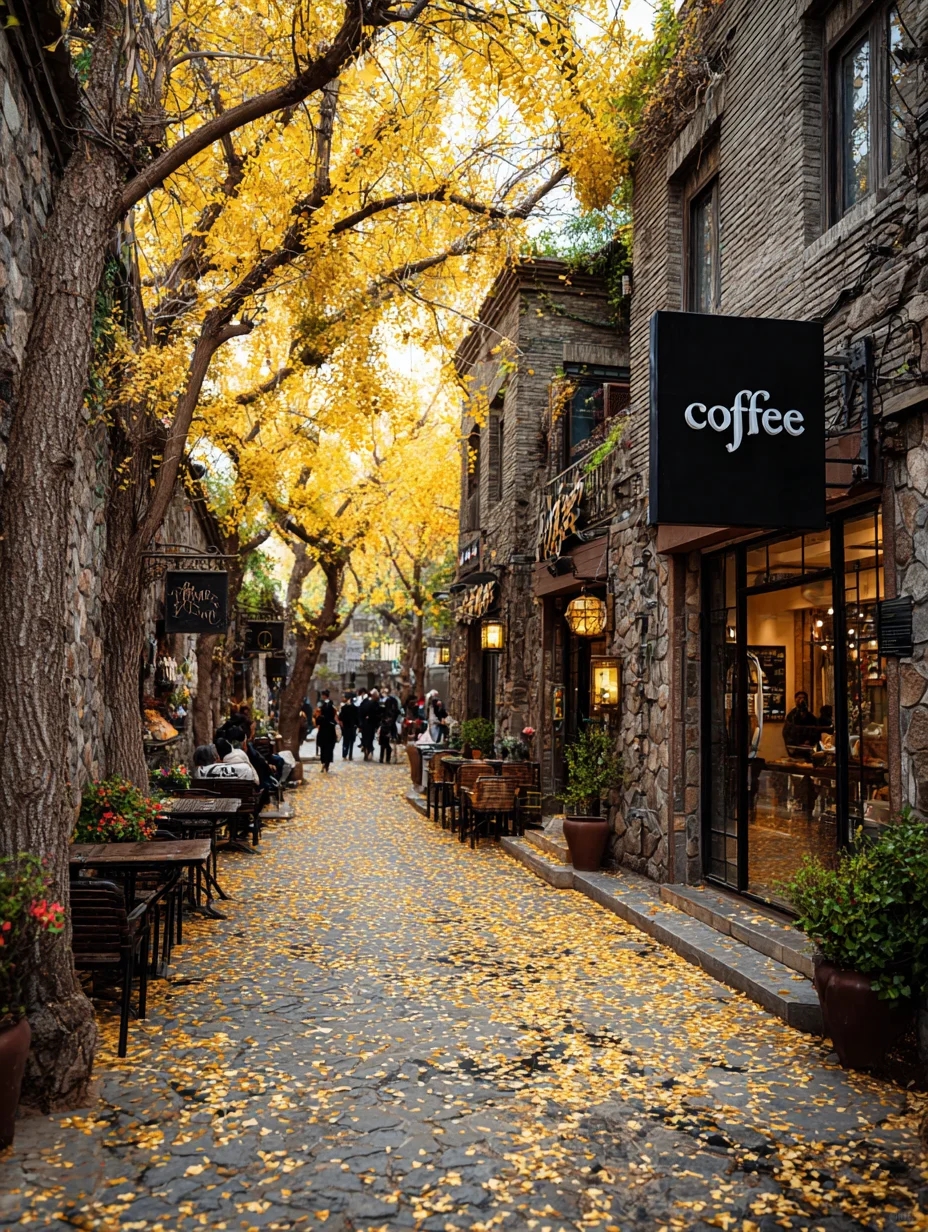}
\hfill
\includegraphics[width=0.24\textwidth]
{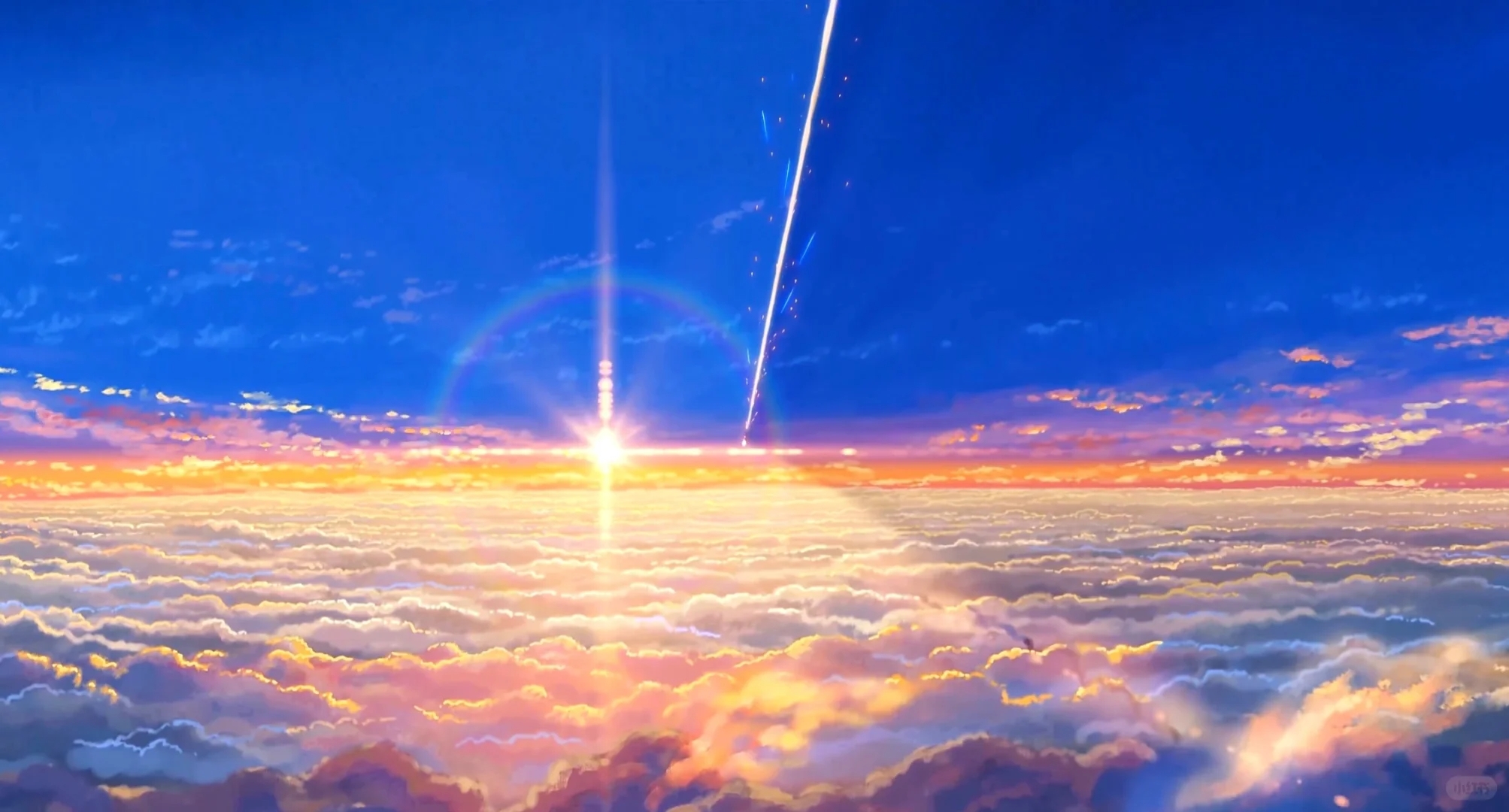}
\hfill
\includegraphics[width=0.24\textwidth]
{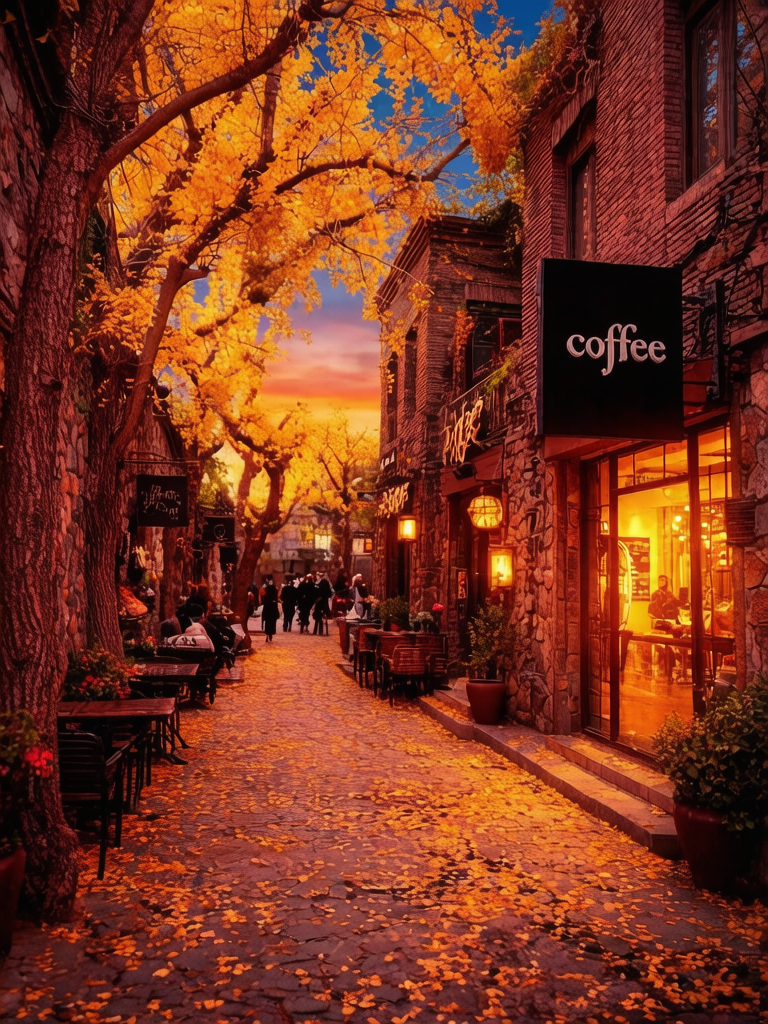}
\hfill
\includegraphics[width=0.24\textwidth]
{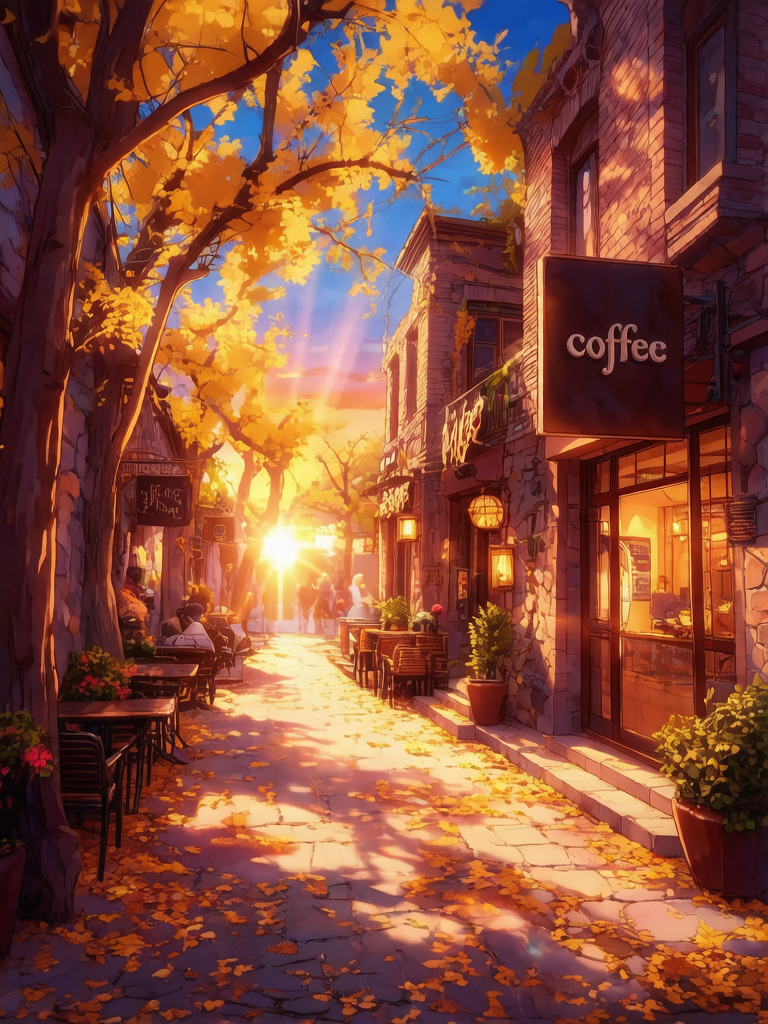}

\vspace{2mm}

{\footnotesize
Let the two puppies run in the snow.
}

\vspace{1mm}

\includegraphics[width=0.24\textwidth]
{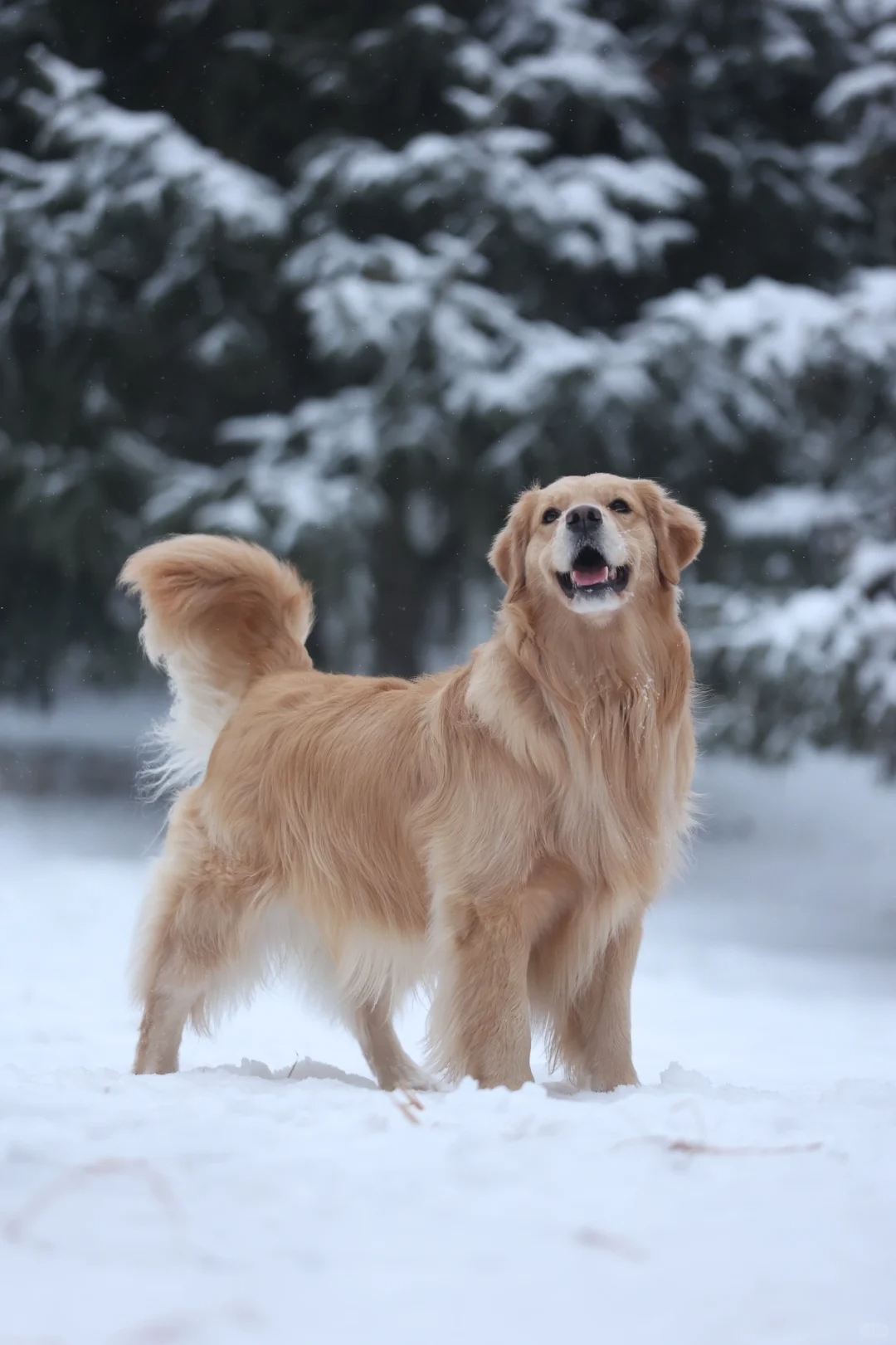}
\hfill
\includegraphics[width=0.24\textwidth]
{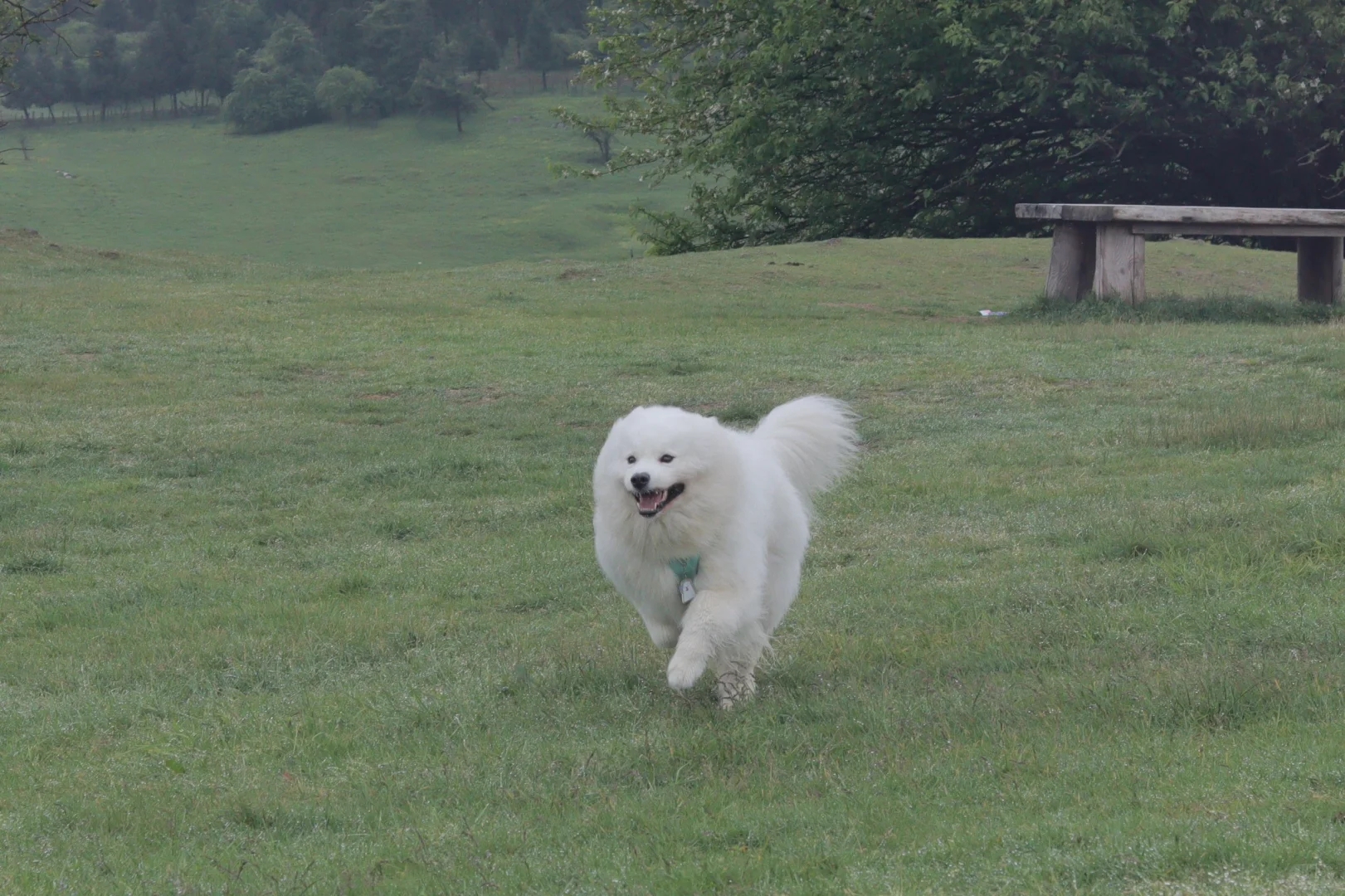}
\hfill
\includegraphics[width=0.24\textwidth]
{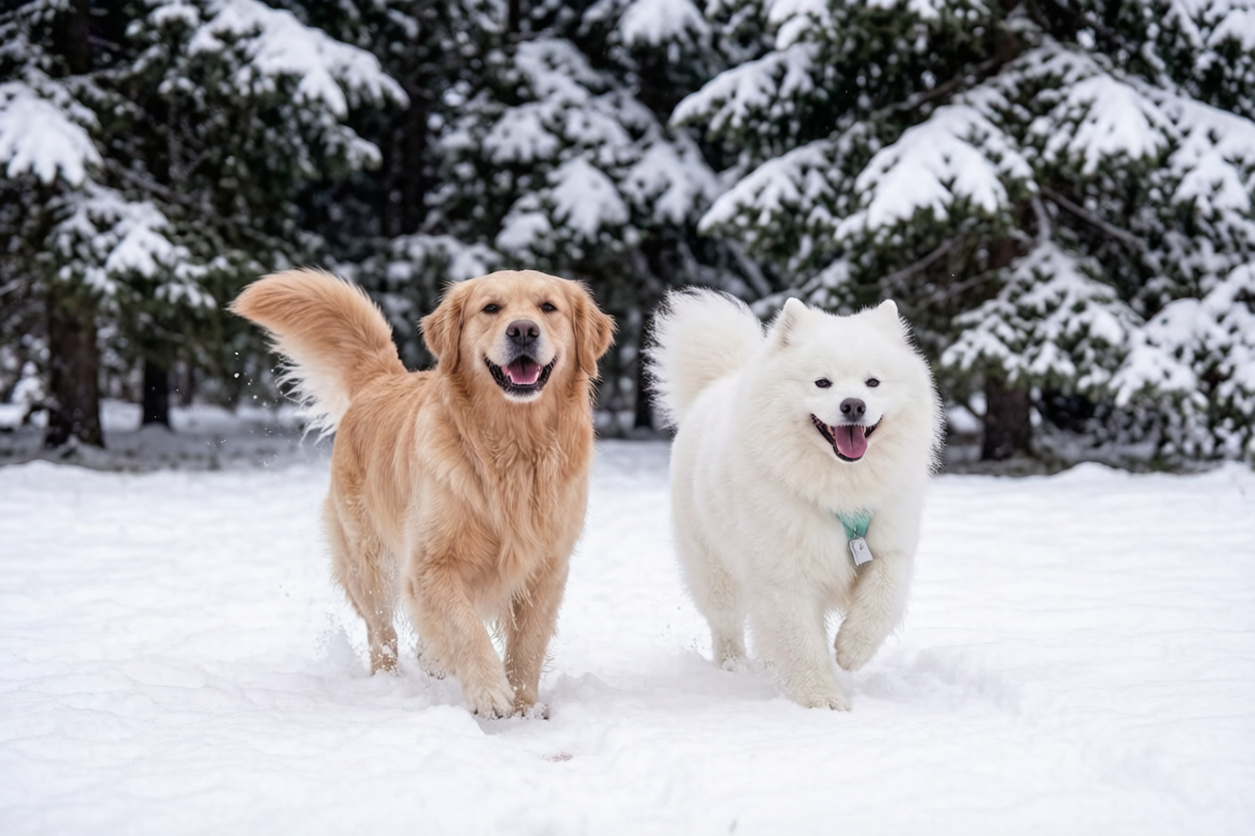}
\hfill
\includegraphics[width=0.24\textwidth]
{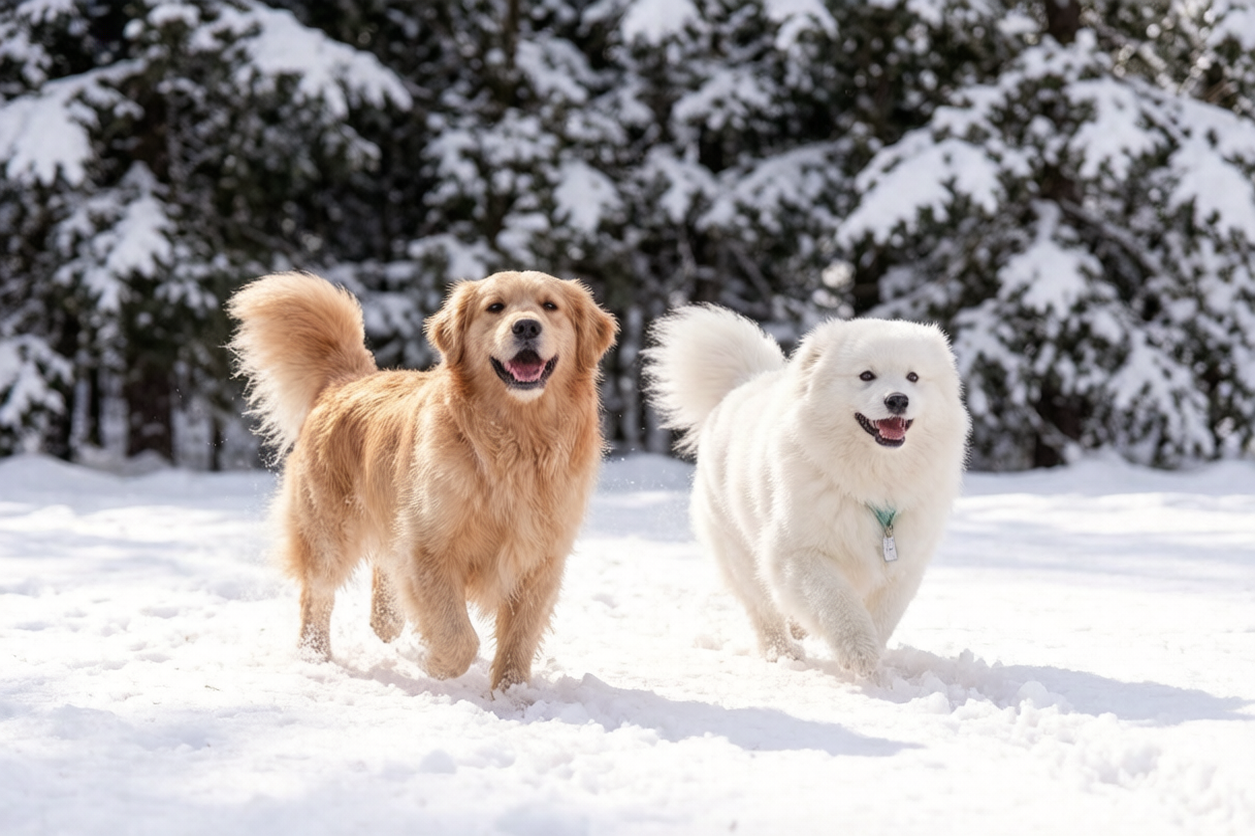}

\vspace{-1mm}

\caption{
Dual-reference image editing examples comparing the original
HunyuanImage-3.0 80B model and our 20B pruned model.
The pruned model preserves multi-image reasoning and
reference-following capabilities despite substantial model compression.
}
\label{fig:dual_ref_case}

\vspace{-3mm}
\end{figure*}
\section{Introduction}

In the past one year, closed-sourced commercial models like Nano-Banana~\citep{nanobanana} and GPT-Image~\citep{gptimage} series dominate the image synthesis scenario. Recently, excellent open-sourced models like HunyuanImage-3.0~\citep{cao2025hunyuanimage} and FLUX.2~\citep{flux2_2025} have been developed to catch them. However, they are normally too costly for practical usage. Typically, HunyuanImage-3.0, albeit achieve SOTA performance among open-sourced models, consume 80B parameters and is recommended to infer at the lowest requirement of 3$\times$80GB GPUs, hindering its popularity in community. To this end, structured pruning is one of the most straightforward methods to alleviate the burden by compressing the original model into a smaller one and is normally hardware-friendly. However, existing structured pruning methods are customized deeply for DiT architecture whereas HunyuanImage-3.0 is built upon an MoE-based decoder-only LLM, \textit{i.e.}, Hunyuan-A13B. The challenge lies in the fact that only an extremely high degree of parameter reduction can make the resulting model lightweight enough to run on low-cost devices. Existing methods could hardly achieve this (normally meet severe performance drop when reducing ratio $\geq 15\%$). In this paper, we propose TMP, a structured pruning framework designed to overcome these challenges. TMP involves a joint pruning strategy, followed by a two-stage post-prune recovery training. The algorithm could be directly applied to the pruning of DiT-based architecture that we generalize to Z-Image turbo without much effort. 

The proposed pruning framework includes a structured pruning followed by a recovery training. The recovery training has two stages: (1) A tree-structured local feature supervision running in divide-and-conquer manner. It is instantiated as a combination of both off-policy and on-policy distillation (OPD). (2) A brief output distillation to match the flow matching velocity prediction. Normally, if network parameters have undergone pruning, each layer-wise inference would inevitably suffer from representation misalignment. In the end-to-end perspective, such errors will accumulate across layers and ultimately lead to model collapse due to cascading error propagation. Preliminary works like \citep{wang2021revisiting} introduce local supervision to mitigate this issue. Basically, tokens processed by a teacher layer are fed into its corresponding succeeding student layer. The supervision is imposed by matching the student outputs against the outputs of the subsequent teacher layer. Recently, some structured pruning works like FastFLUX~\citep{cai2026fastflux} and PPCL~\citep{ma2025pluggable} adopt the local supervision design in their algorithm. Our method, TMP-Pruner differs from theirs in two key aspects, which inspire its name: (a) \textbf{T}ree-structured Merging: We maintain the intervals with a tree-structured merging. Hierarchically, two adjacent intervals (layer sequences) may be merged into a single one, in which case their separate distillation objectives are replaced by a unified loss after merging. This induces a curriculum-learning-like behavior, the model gradually shifts toward newly formed unified objectives as the local supervision regions get progressively merged. Eventually, the objective will reduce to a single end-to-end token representation matching. (b) \textbf{O}n-policy Distillation: We extra introduce an on-policy feature distillation and extend it to a mixed policy distillation that further align the pruned model to the original model. 

In short, our contributions are fourfold:
\begin{itemize}
    \item We propose a unified structured pruning framework that involves a heuristic pruning followed by a mixed policy distillation that refines the teacher-student alignment in intermediate token space.
    \item We devise a novel tree-structured manner of curriculum learning scheme to ease the optimization.
    \item Experiments demonstrate that our proposed framework could be applied to SOTA open-sourced text-to-image/image-editing model at a huge parameter reduction ratio, making it become affordable, with limited losses of generation quality.
    \item Our framework is versatile towards different backbone architectures (MoE and DiT), model sizes (80B HunyuanImage-3.0 and 6B Z-Image turbo) and image tasks (image editing and text-to-image generation).
\end{itemize}

\section{Related Work}

Generative models for high-fidelity image generation have become popular in recent years. They tend to obtain better performance at the cost of increasing model capacity, typically parameter consumption. This imposes enormous memory consumption that either prohibits the model from running on a single consumer-level GPU or leads to low throughput in large-scale deployment. Structured pruning is intuitive to relieve this by slimming the network architecture to obtain a lightweight version. It involves a model pruning stage followed by a recovery training stage. 

\subsection{Structured Pruning} The ways to determine pruned model architecture could be divided into five types: (a) Importance metric (b) Probabilistic modeling (c) NAS-based sampling (d) Cheap replacement and (e) Time-step sensitive allocation. \textbf{(a)} Molchanov et al.~\citep{molchanov2019importance} came up with a layer importance metric to determine which parameters to prune. Diff-Pruning~\citep{fang2305structural} adopted a time-step-aware version. LD-Pruner~\citep{castells2024ld} further proposed an operator-level metric. Xie et al.~\citep{xie2025sana} used cosine similarity given input and output of each layer to measure the importance. OBS-DIFF~\citep{zhu2025obs} invented a second-order sensitivity based on Hessian matrix computation. PPCL~\citep{ma2025pluggable} combined linear probing with CKA-analysis to conduct contiguous layer set removal. \textbf{(b)} EcoDiff~\citep{zhang2024effortless} learned a differentiable mask for sparsification. TinyFusion~\citep{fang2025tinyfusion} proposed a differentiable masked sampling that learned shallow DiT sub-networks to reduce denoising computation through adaptive depth reduction. 
\textbf{(c)} ALTER~\citep{yang2026alter} trained a supernet and used dynamic routing for block and layer pruning. E-DiT~\citep{wang2026elastic} trained differentiable router modules for block skipping and dimension reduction in linear layers. 
\textbf{(d)} Unlike hard pruning, some works conducted cheap substitution of existing modules. FastFLUX~\citep{cai2026fastflux} progressively replaced the DiT blocks as linear modules that will join a sandwich training scheme. Amber-Image~\citep{yang2026amber} proposed a method that could merge two-stream architecture in MMDiT~\citep{esser2024scaling} into a unified stream. However, the sandwich training in FastFLUX demands constructing new local supervision datasets every time when a transformer block get replaced. The two-stream merging strategy in Amber-Image is only specialized for two-stream DiT architecture. \textbf{(e)} Some works introduced a time step-sensitive design related to the reverse diffusion nature during runtime. MosaicDiff~\citep{guo2025mosaicdiff} handcrafted a sparsity curve that set model variants with different sizes under different time steps. Diff-ES~\citep{liu2026diff} automated it by setting the sparsity schedule with evolutionary search. The former requires deploying a series of pruning variants that is unfriendly towards low-memory devices. The latter attains the entire supernet during inference that demands the same memory budget as the original model in the worst-case scenarios.

\subsection{Recovery Training}
After architecture pruning, recovery training will transfer knowledge (KD)~\citep{kd} from the original model to the pruned model. Besides naive KD, some works emphasized on optimizing the loss design. HierarchicalPrune~\citep{kwon2026hierarchicalprune} used reverse loss weighting to suppress the learning of layers with high importance. IGSM~\citep{zheng2025igsm} proposed a second-order Jacobian matching loss inspired by Finite-Time Lyapunov Exponents that makes the model more robust towards input perturbation in the denoising trajectory. However, the loss weighting strategy requires human efforts for delicate adjustment. The second-order metric brings about heavy computation budget. Regardless of the choice of loss function, the optimization process is fundamentally driven by the low-level intermediate feature representations, higher-level latent or noise estimation. These methods, in principle, still exhibit off-policy behavior due to their dependence on teacher-generated trajectories. Recently, in language modeling tasks, on-policy distillation (OPD) has been shown to achieve faster and more effective distillation than off-policy distillation~\citep{opd-tml,agarwal2024policy}. The key is that student rollout affects future supervision distribution. 
In this paper, we devise an on-policy paradigm that constructs supervisory target given student intermediate representations. It runs in mixed mode that alternates off-and-on policy distillation.

Moreover, existing structured pruning works are restricted to text-to-image task with DiT architecture, not involving architecture like MoE adopted by modern SOTA models and image editing task. Our method is able to prune HunyuanImage-3.0 at high parameter reduction ratio. 

\section{Methodology}

\subsection{Structured Pruning}\label{sec:HP} 

HunyuanImage-3.0~\citep{cao2025hunyuanimage} is an 80B MoE-based image editing model built upon Hunyuan-A13B~\citep{hunyuan_a13b} where the MoE layers encompass 98\% of the parameters. We perform a joint expert and width pruning that add up to 75\% of parameter count reduction, resulting in a 20B pruned model. 
(1) \textbf{width pruning}: We apply magnitude pruning to reduce the intermediate size of the MLPs in MoE-FFN from 3072 to 2048. (2) \textbf{expert pruning}: We aggregate the gating scores over a calibration set and keep only 24 out of 64 experts per layer. Z-Image turbo~\citep{cai2025z} is a 6B single-stream DiT-based image generation model. We apply width pruning to it that is highly similar to what we have done above for HunyuanImage-3.0. We reduce the expansion rate of the MLP by $37.5\%$ and achieve an overall 33\% of parameter count reduction. This results in a 4B pruned model.

\subsection{Recovery Training}\label{sec:recovery_training}

The recovery training runs in two-staged fashion. The first stage is about intermediate token distillation. The second stage further refines by velocity prediction distillation. We parameterize the full model as $F_\theta$. It consists of three components: a collection of input encoders and embedders $E_\theta$ (VAE, timestep embedders and any other involved encoders), a transformer backbone $\pi_\theta$ and two parallel prediction heads — a velocity head $v_\theta$ for image latents and an autoregressive head $p_\theta$ for text tokens. Across all training stages, $E_\theta$ is kept frozen. In the mixed-policy feature-distillation stage we unfreeze only $\pi_\theta$; in the subsequent prediction-alignment stage we additionally unfreeze the heads ${v_\theta, p_\theta}$. We generalize $\pi_\theta$ (with a slight abuse of notation) from the next-token distribution to the sequence of intermediate features produced by the student backbone, thereby moving the on-/off-policy distinction from the output layer down to intermediate representations.

\subsubsection{Tree-Structured Mixed-policy Feature Distillation}\label{sec:off_policy_distillation}

Supposed there are total $L$ layers in each model, we set $L$ intervals correspondingly where each interval contains one layer at the beginning. We conduct the distillation fashion in a tree-structured, divide-and-conquer manner.
 
\textbf{Mixed policy distillation}. Supposed there are M intervals in current training iteration, we illustrate the local distillation of interval $i$ without loss of generality. It satisfies $L=\sum_{i=1}^{M} B_i$ where $B_i$ denotes the number of consecutive layers covered by interval $i$ . Supposed interval $i$ begins at layer $l$, we introduce how local feature distillation is conducted between the original model $\Pi_*$ and the pruned model $\Pi_\theta$. In each iteration, we feed the token sequence $x \sim p_{\mathrm{data}}$ into the teacher model to perform end-to-end forward propagation. In this procedure, we collect input tokens into layer $l$, said $X_*^{l-1}$ and the output tokens by layer $l+B_i-1$, said $X_*^{l+B_{i}-1} \equiv \Pi_{*}^{1 \rightarrow l+B_i-1}(x)$.
For student, instead of regular forwarding, we feed the input tokens into interval $i$ of teacher into that of the student to get the output tokens. We make it learn towards corresponding teacher output tokens, forming off-policy distillation:

\begin{equation}
\label{eq:offpolicydistill}
\mathcal{L}_{\mathrm{off}}^{(i)}
=
\min_{\theta}
\mathbb{E}_{x \sim p_{\mathrm{data}}, X_{*}^{l-1} \sim \Pi_{*}^{1 \rightarrow l-1}(x)}
\left[
\frac{1}{N}
\mathcal{D}
\left(
X_*^{l+B_{i}-1},
\Pi_{\theta}^{\,l \rightarrow l+B_i-1}(X_{*}^{\,l-1})
\right)
\right]
\end{equation}

where $N$ is the number of tokens. $\mathcal{D}(\cdot, \cdot)$ represents the feature discrepancy. We instantiate it as L2 distance. Besides the off-policy distillation, we enhance the intermediate token representation learning towards the original model by introducing on-policy distillation to the feature distillation. 
We instantiate the on-policy feature distillation by swapping the roles described above. The key lies in utilizing the student roll-outs to join the learning target construction. 
We condition on the imperfect student representation and modulate it by local original model consecutive layers to construct learning target and define the distillation as:
 
\begin{equation}
\label{eq:onpolicydistill}
\mathcal{L}_{\mathrm{on}}^{(i)}
=
\min_{\theta}
\mathbb{E}_{x \sim p_{\mathrm{data}}, X_{\theta}^{l-1} \sim \Pi_{\theta}^{1 \rightarrow l-1}(x)}
\left[
\frac{1}{N}
\mathcal{D}
\left(
\mathrm{sg}
\left[
\Pi_{*}^{\,l \rightarrow l+B_i-1}(X_{\theta}^{l-1})\right],
X_{\theta}^{l+B_{i}-1}
\right)
\right]
\end{equation}
where $\mathrm{sg}\left[\cdot\right]$ denotes the stop-gradient operator. We combine both mimicking loss, leading to a mixed policy distillation manner. The ultimate learning objective considering all current $M$ intervals is $\mathcal{L}_{\mathrm{mixed}}=\frac{1}{M}\Sigma_{i=1}^{M}\left[\mathcal{L}_{off}^{(i)}+\mathcal{L}_{on}^{(i)}\right]$

 \textbf{Tree-structured Interval Merging}. To gradually mitigate the propagated error towards end-to-end optimization, we adopt a bottom-up binary tree fashion of interval merging. As every $K$ training iterations go by, two adjacent intervals are merged into one, halving the interval number.

\subsubsection{Prediction Alignment}
After hierarchical mixed policy feature distillation, we use output distillation\footnote{For HunyuanImage-3.0 which demands self-recaption capability, we extra apply a Kullback-Leibler (KL) divergence between the next-token prediction by the language modeling heads of both models}\label{sec:prediction_distillation} to further align the original model and the pruned model. It is accomplished by distilling the velocity prediction between both models as shown in Eq.~\ref{eq:velocity_distill}. 

\begin{equation}
\label{eq:velocity_distill}
\mathcal{L}_{\mathrm{velocity}}
=
\mathbb{E}_{x_t,t,c}
\left[
\left\|
v_{\theta}(x_t,t,c)
-
v_{*}(x_t,t,c)
\right\|_2^2
\right]
\end{equation}

\section{Experiments}

\begin{figure}[htbp]
    \centering
    \includegraphics[width=0.50\linewidth]{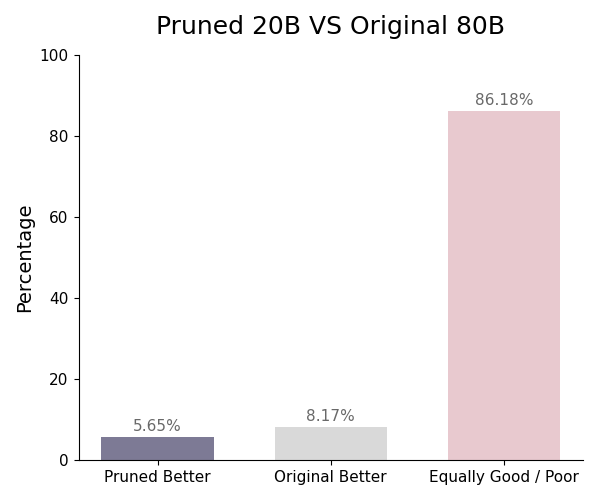}
    \caption{Human preference comparison between pruned 20B \textit{v.s.} original HunyuanImage-3.0 80B on private image editing benchmark. The evaluation shows a limited performance degradation ($-2.5\%$) on human preference.}
    \label{fig:pruned_vs_original_gsb}
\end{figure}

\subsection{About HunyuanImage-3.0}\label{exp:hunyuanimage}

(a) \textbf{Original model \textit{v.s.} pruned model}
Fig.~\ref{fig:pruned_vs_original_gsb} shows the comparison between HunyuanImage-3.0 original 80B model and its pruned 20B model by our proposed structured pruning method. Notably, there is only -2.5\% of human preference dropping. The comparison is made upon private image editing dataset. Fig.~\ref{fig:single_ref_case} and Fig.~\ref{fig:dual_ref_case} have shown some image editing results comparison between these two models.

(b) \textbf{Memory-Efficient Inference}
To enable deployment on consumer-grade GPUs, we develop a memory-efficient inference framework based on FP8 quantization and dynamic module offloading. The main model is quantized to FP8 precision, significantly reducing the memory footprint. To avoid memory spikes during initialization, model weights are first loaded into CPU memory and selectively transferred to the GPU according to the execution stage. In addition, several memory-intensive components, including the visual encoder and VAE, are activated on demand and immediately offloaded after use. Intermediate activations and CUDA caches are also released between major generation stages to minimize memory fragmentation and peak allocation. With these optimizations, our model can perform 1024$\times$1024 image generation with 8 sampling steps on a single NVIDIA RTX 4090 (24 GB), requiring less than 24 GB of GPU memory at peak usage.

\begin{table}[t]
\centering
\caption{Quantitative evaluation results on OneIG-EN.}
\label{tab:oneig_en}
\begin{tabular}{lcccccc}
\toprule
Model & Alignment & Text & Reasoning & Style & Diversity & Overall \\
\midrule

Qwen-Image 20B \,\cite{wu2025qwen}
& \textbf{0.882} & 0.891 & \textbf{0.306} & \textbf{0.418} & \textbf{0.197} & 0.539 \\

Z-Image 6B\,\cite{cai2025z}
& 0.881 & 0.987 & 0.280 & 0.387 & 0.194 & \textbf{0.546} \\

Z-Image-Turbo 6B\,\cite{cai2025z}
& 0.840 & \textbf{0.994} & 0.298 & 0.368 & 0.139 & 0.528 \\

\midrule

PPCL-OPPO-10B\,\cite{ma2025pluggable}
& 0.839 & 0.860 & 0.249 & 0.359 & 0.121 & 0.485 \\

Amber-Image-10B\,\cite{yang2026amber}
& \textbf{0.867} & 0.938 & 0.278 & 0.298 & 0.137 & 0.504 \\

Amber-Image-6B\,\cite{yang2026amber}
& 0.829 & 0.917 & 0.284 & 0.287 & 0.135 & 0.490 \\

Z-Image-Turbo pruned-4B (Ours)
& 0.840 & \textbf{0.980} & \textbf{0.305} & \textbf{0.364}  & \textbf{0.161} & \textbf{0.530} \\
\bottomrule
\end{tabular}
\end{table}

\begin{table}[t]
\centering
\caption{Quantitative evaluation results on OneIG-ZH.}
\label{tab:oneig_zh}
\begin{tabular}{lcccccc}
\toprule
Model & Alignment & Text & Reasoning & Style & Diversity & Overall \\
\midrule

Qwen-Image 20B\,\cite{wu2025qwen}
& \textbf{0.825} & 0.963 & 0.267 & \textbf{0.405} & \textbf{0.279} & \textbf{0.548} \\

Z-Image 6B\,\cite{cai2025z}
& 0.793 & \textbf{0.988} & 0.266 & 0.386 & 0.243 & 0.535 \\

Z-Image-Turbo 6B\,\cite{cai2025z}
& 0.782 & 0.982 & \textbf{0.276} & 0.361 & 0.134 & 0.507 \\

\midrule

PPCL-OPPO-10B\,\cite{ma2025pluggable}
& \textbf{0.854} & 0.878 & \textbf{0.268} & \textbf{0.365} & 0.130 & 0.499 \\

Amber-Image-10B\,\cite{yang2026amber}
& 0.798 & \textbf{0.975} & 0.221 & 0.362 & 0.153 & \textbf{0.502} \\

Amber-Image-6B\,\cite{yang2026amber}
& 0.779 & 0.953 & 0.208 & 0.345 & 0.143 & 0.486 \\

Z-Image-Turbo pruned-4B (Ours)
& 0.784 & 0.946 & 0.261 & 0.352 & \textbf{0.158} & 0.500 \\

\bottomrule
\end{tabular}
\end{table}

\begin{table}[!t]
\centering
\caption{Quantitative evaluation results on LongText-Bench.}
\label{tab:longtext_bench}
\begin{tabular}{lcc}
\toprule
Model & LongText-Bench-EN & LongText-Bench-ZH \\
\midrule

Qwen-Image 20B\,\cite{wu2025qwen}
& \textbf{0.943} & \textbf{0.946} \\

Z-Image\,\cite{cai2025z}
& 0.935 & 0.936 \\

Z-Image-Turbo\,\cite{cai2025z}
& 0.917 & 0.926 \\

\midrule

PPCL-OPPO-10B\,\cite{ma2025pluggable}
& 0.871 & 0.885 \\

Amber-Image-10B\,\cite{yang2026amber}
& \textbf{0.911} & \textbf{0.915} \\

Amber-Image-6B\,\cite{yang2026amber}
& 0.870 & 0.876 \\

Z-Image-Turbo pruned-4B (Ours)
& 0.889 & 0.882\\
\bottomrule
\end{tabular}
\end{table}

\subsection{About Z-Image turbo}\label{exp:zimage}
Both PPCL-OPPO and Amber-Image are among the best T2I models obtained via structured pruning in recent literature. They are built upon Qwen-Image (better than our experimenting teacher Z-Image-Turbo in the metrics shown in below involved benchmarks). Table~\ref{tab:oneig_en}, Table~\ref{tab:oneig_zh} and Table~\ref{tab:longtext_bench} show the comparison of the pruned Z-Image Turbo 4B model by our method to others on OneIG-ZH/EN\,\cite{chang2025oneig} and LongText benchmarks\,\cite{geng2025x} respectively. Notably, the 4B-pruned model outperforms the pruned models by these methods (OneIG-EN) or is comparable (OneIG-ZH, LongText-Bench) to them despite containing 1.5 to 2.5 $\times$ fewer parameters.

\section{Conclusion}
We presented TMP, a structured pruning framework engineered to compress modern, large-scale image synthesis models. By advancing beyond standard distillation with a tree-structured merging strategy and mixed-policy feature alignment, the proposed framework effectively scales to intricate MoE-based and DiT architectures. TMP achieves an aggressive compression ratio—most notably reducing HunyuanImage-3.0 from 80B to 20B parameters—while preserving high-fidelity generation quality. This work bridges the gap between state-of-the-art generative capabilities and resource-constrained hardware deployment. Moving forward, we aim to adapt this compression paradigm to other generative modalities and investigate deeper hardware-level optimization.

\newpage 
\bibliography{iclr2025_conference}
\bibliographystyle{iclr2025_conference}

\end{document}